\documentclass[runningheads]{llncs}
\usepackage[T1]{fontenc}
\usepackage{graphicx}
\usepackage{amsmath}
\usepackage{latexsym}
\usepackage{amssymb}
\usepackage{booktabs}
\usepackage{enumitem}
\usepackage{graphicx}
\usepackage{color}

\usepackage{comment} 
\usepackage{mathtools} 
\usepackage{tablefootnote} 

\begin{document}
\title{Bidirectional Awareness Induction in Autoregressive Seq2Seq Models}
%
%
\author{Jia Cheng Hu\inst{1}\orcidID{0009-0008-1611-966X} \and
Roberto Cavicchioli\inst{1}\orcidID{0000-0003-0166-0898} \and
Alessandro Capotondi\inst{1}\orcidID{0000-0001-8705-0761}}
\authorrunning{J. C. Hu et al.}
%
\institute{
University of Modena and Reggio Emilia, via G.Campi 213/b, 41125, Modena, Italy \email{name.surname@unimore.it}}
\maketitle              

\begin{abstract}
Autoregressive Sequence-To-Sequence models are the foundation of many Deep Learning achievements in major research fields such as Vision and Natural Language Processing. Despite that, they still present significant limitations. For instance, when errors occur in the early steps of the prediction, the whole output is severely affected. Such reliance on previously predicted tokens and the inherent computational unfriendliness of sequential algorithms, motivated researchers to explore different architectures and methods in the search for bidirectional approaches. In this work, we introduce the Bidirectional Awareness Induction (BAI), a training method that leverages a subset of elements in the network, the Pivots, to perform bidirectional learning without breaking the autoregressive constraints. To showcase its flexibility, we apply the method to three architectures, the Transformer, ExpansionNet v2 and GPT, then perform experiments over three tasks. Experimental results showcase BAI's effectiveness on all selected tasks and architectures. In particular, we observed an increase of up to 2.4 CIDEr in Image-Captioning, 4.96 BLEU in Neural Machine Translation, and 1.16 ROUGE in Text Summarization compared to the respective baselines. Notably, BAI not only has a positive impact on models trained from scratch but on pre-trained models as well. Such an aspect, combined with the absence of architectural requirements synergizes well with the current trend of LLMs. 



\end{abstract}

\section{Introduction}
\label{sec:introduction}

Many tasks in Natural Language Processing (NLP) such as
Neural Machine Translation (NMT) \cite{zhang2018asynchronous,gu2020fully}, Text Summarization (TS) \cite{chung2022scaling,chen-etal-2021-dialogsum,Kim:2019:NAACL-HLT} and Image Captioning (IC) \cite{vinyals2015show,hu2023exploiting} deal with the challenging task of generating meaningful and linguistically correct sentences. This is commonly accomplished by Neural Networks. Typically, models follow the Autoregressive property, meaning that the token distribution predicted on time step $t$, depends on all the previous tokens from 1 to $t-1$. While this approach is intuitive, as the sequential process resembles on a superficial level, how humans communicate, it poorly reflects how we process information, and presents in fact, some limitations. 
If errors in previous predictions occur, the quality of subsequent predictions is undermined. Additionally, unidirectional decoding fails to capture bidirectional contexts that can be exploited for more effective learning. 

Several works in Natural Language Processing (NLP) related fields such as Image Captioning (IC) and Neural Machine Translation (NMT) proposed several approaches to combat the limitations of unidirectional decoding. 
Existing methods can be categorized into two main non-exclusive classes we name for simplicity "architecture-based" and "algorithmic-based". The first \cite{zhang2019regularizing,zhang2018asynchronous,zhou2019synchronous,zhou2022compact} consists of feeding Right-to-Left (R2L) data (or processing the input in a reversed order) to the network, in addition to the standard Left-to-Right (L2R) which often imply also architectural modifications. These methods lead to better performances at the expense of a higher computational cost. The second consists of training and algorithmic modifications \cite{wang2018semi,gu2017non,mehri2018middle,sun2017bidirectional} which do not focus on the architecture but propose a different framework to predict multiple tokens simultaneously, often at the cost of the final accuracy. In this category, fall the very recent works of Text Diffusion models \cite{liu2024text,li2022diffusion}.
These methods focus on predicting the result in one single or multiple parallel passages in contrast to the standard autoregressive models.

Architecture-based strategies are very effective in integrating the R2L processing in the model but typically require modifications in the architecture, which means that running the model is typically more time-consuming and cannot be easily extended to Large Pre-trained Models. The opposite was generally true (with some exceptions \cite{sun2017bidirectional}) in the case of algorithm-based strategies where lowering the inference cost was the most desirable effect at the expense of a negligible accuracy degradation. Finally, Text Diffusion models seem to be a promising direction which offers both low latency and satisfying output quality. However, they still present limiting factors, such as a high test-training discrepancy \cite{tang2023diffusion} or the dependency on additional components such as length classifiers \cite{liu2024text} and, overall, autoregressive decoding still represents the most popular and solid approach in modern applications. 

In this work, we introduce a proposal that shares the same purpose as architectural-based methods and aims at achieving better performance. However, it operates only during the training and does not require architecture modifications, which can be ideal in the case of pre-trained models \cite{radford2019language,devlin2018bert,chung2022scaling}. In particular, we observe that not all parts of a typical encoder-decoder network are subjected to the auto-regressive constraint and introduce the concept of \textit{Pivots}. That is, there are elements in the network that are allowed to access and be trained on the entire target sequence, and we leverage them to induce bidirectional awareness in the network.


Overall, the paper is organized as follows. First, we introduce the concept of pivots and Bidirectional Awareness Induction training. We showcase its application in three architectural instances. Then, we describe the experimental setup and showcase the results in three tasks. Afterwards, we compare BAI with other methods and analyze the quality of pivots. Finally, we discuss the limitations and present the conclusion and future works. The contributions are the following:
\begin{enumerate}
\item We introduce the concept of \textit{Pivots}, described as network elements on which we perform training on tasks that can be beneficial but are not directly related, to the final objective;
\item Leveraging the concept of pivots, we introduce the \textit{Bidirectional Awareness Induction} strategy, which trains pivots on a bidirectional loss without breaking the autoregressive property of the model, preserving the advantages of the two worlds.
\item We showcase our method's flexibility and robustness over various architectures, tasks and training setups. Notably, our method works on both pre-trained models and models that are trained from scratch.
\end{enumerate}
\section{Related Works}
\label{sec:related}

Several approaches have been proposed over the past years to combat the limitations of unidirectional decoding, mostly in NMT and IC. Since the decoding stage in these two applications is similar and solutions are often interchangeable, in this Section we report related works in both fields. 

In NMT, the works of \cite{liu2016agreement}
and \cite{zhang2019regularizing} 
trained two models for the L2R and the R2L decoding respectively and combined their results during inference time they proposed the joint search, an alternative to the beam search. 
\cite{sun2017bidirectional} 
proposed the Bidirectional Beam Search 
\cite{wang2018semi} 
implemented a Semi-Autoregressive architecture that decodes multiple tokens at each step.
\cite{zheng2018modeling}
proposed separated layers for the past and future representations in an RNN-based decoder. 
In the work of \cite{mehri2018middle} 
they propose an alternative decoding order that starts from the middle of the sequence, in contrast to the standard L2R and R2L decoding. Whereas,
\cite{zhou2019sequence} 
completed the idea using the complementary approach of ordering the decoding stage from the sides to the middle.
\cite{zhang2018asynchronous,tan2020neural} 
introduced the "Asynchronous" bidirectional decoding based on two RNNs, trained for L2R and R2L decoding respectively. First, one decoder produces the R2L sequence, then the second RNN leverages the first result during the generation of the L2R sequence.
\cite{zhou2019synchronous} 
in contrast, introduces the "Synchronous" version, in which both L2R and R2L are generated simultaneously. 
Another notable approach is represented by NAT \cite{gu2017non,gu2020fully,ding2021improving} 
the Non-Autoregressive Architecture whose main principle consists of replicating the source embeddings several times (according to the so-called "fertility") in the decoder, so the latter can perform parallel and bidirectional processing. 

In IC, the work of \cite{wang2018image} 
first proposed the adoption of Bi-LSTM in the field. In  \cite{sammani2019look} 
and \cite{sammani2020show} 
the authors adopted an auxiliary network to perform editing operations on the final result, which mitigates the limitations of the autoregressive decoding.
In CAAG \cite{song2021image} 
the authors propose two models, a primary network that generates a caption greedily. A second one that leverages the first prediction to look at both the past and future and perform a joint beam search. Both predictions are then combined in the generation of the final description.
In CBTIC \cite{zhou2022compact} 
the authors augment the Transformer decoder by designing a particular architecture to integrate also the R2L data in the network.  

Text Diffusion models \cite{liu2024text,li2022diffusion} 
represent a recent and promising emerging family diffusion-based generative models for text. They are capable of performing bidirectional processing and prediction, however, they currently suffer from non-negligible issues,  such as those related to the test-train discrepancy. Our work is orthogonal to these approaches and focuses on autoregressive models, still being the predominant methodology in sequence-to-sequence problems.

\begin{figure*}[h]
  \centering
  \includegraphics[width=0.95\textwidth]{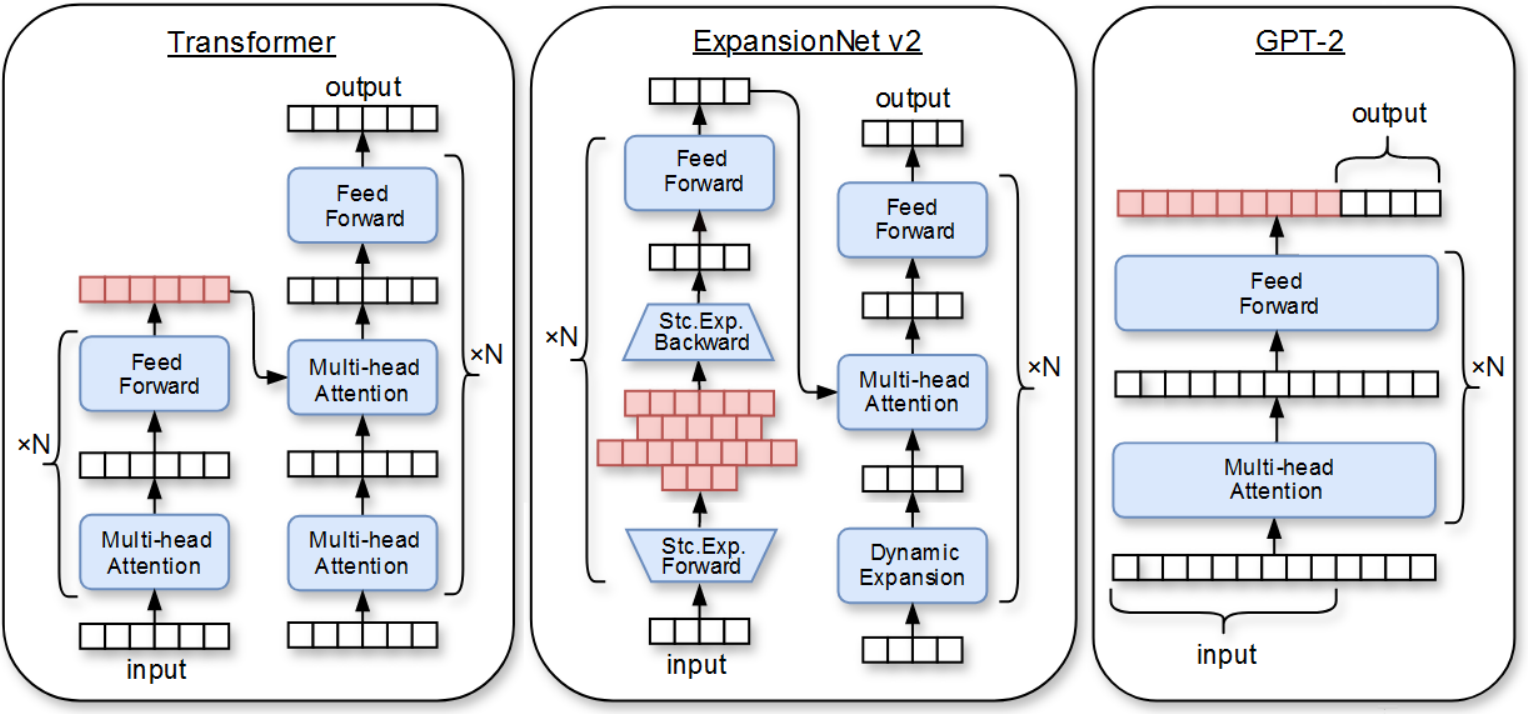}
  \caption{ \label{figure:arch} Pivot selection in case of Transformer \cite{vaswani2017attention}, ExpansionNet v2 \cite{hu2023exploiting} and GPT-2 \cite{radford2019language} architectures. Pivots are highlighted in red colour, processing layers are depicted in blue. 
  }
\end{figure*}

\section{Method}

In this Section, we present the concept of Pivot and the main aspects of our Bidirectional Awareness Induction (BAI), from a conceptual point of view. Then, we apply it to several architectural instances.
 
\subsection{Bidirectional Awareness Induction (BAI)}

Our method, called Bidirectional Awareness Induction (BAI), is based on the concept of Pivots elements. We define Pivots as elements of the network that can be trained on tasks that are not necessarily correlated to the final objective function but can be beneficial to improve the quality of the result. 
In this work, we train pivots to reproduce the target output in Seq2Seq problems and they are selected such that the auto-regressive property of the network is preserved. In this way, we induce bidirectional awareness in auto-regressive models or relax the prediction dependency on the previously generated tokens. Ultimately, BAI  intends to combine the best of two worlds without inheriting the obstacles of bidirectional models.
Overall, BAI can be broken down into three steps: \emph{(i) Pivot Selection}. Select a set of elements to be trained on the bidirectional task (e.g. encoder features). 
\emph{(ii) Length Equalization}. Leverage decoder representations only to equalize the pivot elements to the one of the decoder.  \emph{(iii) Decoder sequence reconstruction}. Reconstruct the decoder sequence using exclusively the result of the previous step. Train the model to optimize the reconstruction discrepancy and the task-specific auto-regressive loss jointly.
The encoder features typically represent the most straightforward selection of pivots since they can be leveraged in the learning without breaking the auto-regression condition. However, the concrete implementation of the strategy depends on the adopted architecture, but notably, BAI does not require architectural modifications. 

We highlighted in the third step that the BAI loss function is intended to be jointly optimized with the Cross-Entropy training. In this way, the network can benefit from an increased bidirectional awareness, without harming the effectiveness of the traditional approach.

In our experiments, to showcase the flexibility of BAI we present in the following sections the application of our proposal to different architectures, such as the popular and established Transformer \cite{vaswani2017attention},  the recent ExpansionNet v2 \cite{hu2023exploiting}, and GPT-2 \cite{radford2019language}. To study the robustness of the idea we performed experiments in multiple tasks such as Neural Machine Translation, Text Summarization, and Image Captioning.

\subsection{BAI in Transformer}
\label{sec:bai_transformer_training}

In this Section, we propose a concrete implementation of the previous idea in the case of the Transformer \cite{vaswani2017attention}. The Transformer is a popular Encoder-Decoder architecture that succeeded in numerous NLP tasks. In essence, all layers are made of Self-Attention and FeedForward layers and their exact implementation is omitted since they do not impact the discussion.

We select the output of the last encoder layer $\overline{E}$=$\{e_1, e_2, \ldots, e_N\}$ to be the \textit{pivots}, which consists of $N$ feature vectors of size $H$. During the second step, the \textit{length equalization} step, we perform the dot product similarity between $\overline{E}$ and the target sequence embeddings $D \in \mathbb{R}^{M \times H}$ where $M$ is the target length. Then apply Softmax and multiply the result by $\overline{E}$:
\begin{equation}
R = Softmax(\frac{(D\overline{E}^{\intercal})}{\sqrt{H}}) \overline{E}
\label{eq:bai_transformer_step2}
\end{equation}
Note that although it involves the embeddings $D$ the auto-regressive property is preserved. The final BAI loss is defined as the Mean Square Loss (MSE) between the results of Equation \ref{eq:bai_transformer_step2} and $D$:
\begin{equation}
    \beta(R, D) = \frac{1}{M}\sum_{t}^{M} \frac{1}{H}(r_t-d_t)^{\intercal}(r_t-d_t)
\label{eq:mse_bai_loss}
\end{equation}
where $r_t \in \mathbb{R}^{H \times 1}$ denotes the element $t$-th element of $R \in \mathbb{R}^{N \times H}$, and $d_t \in \mathbb{R}^{H \times 1}$ defines the $t$-th target embedding. During the training, Equation \ref{eq:mse_bai_loss} is jointly trained with the standard Cross-Entropy (CE):
\begin{equation}
CE+BAI(Y,X) = \lambda \beta(Embed(Y),X) + \sum_{t=1}^{|M|} log \ p(y_t|y_{i<t}, X)
\label{eq:bai_augmented_xe}
\end{equation}
where $X$ and $Y$ denote the input and target sequences respectively. $Embed(\cdot)$ denote the embedding function, and $\lambda$ is a configurable hyper-parameter.

\subsection{BAI in ExpansionNet v2}
\label{sec:bai_expansionnet_training}

ExpansionNet v2 \cite{hu2023exploiting} is another Encoder-Decoder architecture developed for Image Captioning. The encoder implements the (Block) Static Expansion, which distributes the input content across a group of arbitrary numbers of vectors, denoted by $G=\{g_1, g_2 \ldots, g_{|G|}\}$, in the Forward Expansion. The result is called "expanded sequence" and is processed again in the Backward Expansion to retrieve the original sequence length. We refer to the original paper for the architectural details. 

In the BAI's Pivot selection step, instead of adopting directly the last encoder outputs, we adopt two intermediate results. Let $A^{g}_{l}, B^{g}_{l} \in \mathbb{R}^{g \times H}$ for $g \in G$, the results of the Forward Expansion in the two operation paths, where $g \in G$ denotes the expanded sequence length and $l$ identifies the layer. In particular, we compute $\overline{A}^{g}=\sum_{l}^{L} {A}^{g}_{l}$ and $\overline{B}^{g}_{l}=\sum_{l}^{L} {B}^{g}_{l}$.

In the \textit{length equalization step} we replicate a parameter-less version of the Backward Expansion:
\begin{equation}
\normalsize
\begin{aligned}
R^{g}_{1} &= \phi(ReLU(\frac{(D((\overline{A}^g+\overline{B}^g)/2)^{\intercal})}{\sqrt{H}})) \ \ \ \ \ for \ g \in G \\
R^{g}_{2} &= \phi(ReLU(-\frac{(D((\overline{A}^g+\overline{B}^g)/2)^{\intercal})}{\sqrt{H}})) \ \ \ \ \ for \ g \in G \\
R_{1} &= [R_{1}^{g_{1}}, R_{1}^{g_{2}}, \ldots, R_{1}^{g_{|G|}}]_{2}, \ \ 
R_{2} = [R_{2}^{g_{1}}, R_{2}^{g_{2}}, \ldots, R_{2}^{g_{|G|}}]_{2} \\
\hat{A} &= [\overline{A}^{g_{1}}, \overline{A}^{g_{2}}, \ldots, \overline{A}^{g_{|G|}}]_{1}, \ \ 
\hat{B} = [\overline{B}^{g_{1}}, \overline{B}^{g_{2}}, \ldots, \overline{B}^{g_{|G|}}]_{1} \\
R &= (\frac{R_1\hat{A}}{|G|} + \frac{R_2\hat{B}}{|G|} )/2 \\
\end{aligned}
\label{eq:bai_step_2_expansionnet}
\end{equation}
where $\phi$ denotes a row-wise normalization function \cite{hu2023exploiting} and $[ \bullet, \bullet, \ldots, \bullet]_{n}$ denotes the concatenation over the $n$-th axis.

The third step of $BAI$ is equivalent to the Equation \ref{eq:mse_bai_loss} the case described in Section \ref{sec:bai_transformer_training}. We define the BAI loss as the MSE between the result of Equation \ref{eq:bai_step_2_expansionnet} and $D$ and perform joint optimization with the Cross-Enropy.

\subsection{BAI in Large Pre-Trained Models}
\label{sec:bai_gpt2_training}

Most popular large pre-trained models are based on the Transformer architecture, such as Flan-T5 \cite{chung2022scaling}, mBART \cite{liu2020multilingual}, and GPT-2 \cite{radford2019language}. 
For this reason, BAI can be applied in the same way as described in Section \ref{sec:bai_transformer_training} with few exceptions.

In the case of GPT-2, the architecture differs compared to the vanilla Transformer because of the absence of the encoder. In this case, we select the Pivot elements to be the final state of the input sequence. 

\section{Experimental Setup}

To evaluate the generalizability of our method we evaluate BAI across three tasks and five architectures. For Neural Machine Translation (NMT) we adopt the Transformer \cite{vaswani2017attention} and mBART \cite{liu2020multilingual}. For Image Captioning (IC) we select the \cite{hu2023exploiting} and \cite{vaswani2017attention}. Finally, Flan-T-Small \cite{chung2022scaling} and GPT-2 \cite{radford2019language} are evaluated on Text Summarization (TS).
The architecture of mBART, Flan-T5-Small, and GPT-2 are pre-trained according to the dataset and modalities reported in the respective works.

\subsection{Dataset}

A total of four datasets are adopted in the experiments. IC is evaluated on the Microsoft's COCO 2014 \cite{lin2014microsoft} which is split according to Karpathy \cite{karpathy2015deep}. Overall, it consists of 113K training images, 5K images for the validation, and an additional 5K for the test. Each image is paired with five ground-truth captions. They are pre-processed by lower casing and punctuation removal. Words that occur less than five times are discarded, resulting in a total of 10K unique tokens.  NMT is tested on the IWSLT 2015 English-Vietnamese (En-Vi) corpus, which consists of 133K sentences for training and 1268 for evaluation. Sequences whose post-tokenization length is greater than 150 are discarded. In all training instances, the target and the source language vocabulary are shared. Each vocabulary is created using the BPE algorithm \cite{sennrich2015neural}. TS is evaluated on the novel TIFU dataset \cite{Kim:2019:NAACL-HLT}, which consists of threads from the "TIFU" subreddit equipped with the "TL;DR" summary. We follow the Pegasus Split \cite{zhang2019pegasus}, the training set consists of $\sim$33K pairs, and 5K are reserved for testing.  Additionally, we adopt the popular DialogSum \cite{chen-etal-2021-dialogsum}, made of 12,460 training and 1500 testing dialogue-summary pairs. Since pre-trained models are adopted for these tasks, no additional pre-processing is applied to the dataset besides the subword tokenization defined in each respective model. The reason we adopted two datasets is motivated by the fact that, in early experiments, some pre-trained baseline models produced only trivial answers in the case of the TIFU dataset. As a result, they were unsuitable for experimentation and comparison. In contrast, all selected models achieved a satisfactory level of performance in the case of DialogSum. In both datasets, training, and testing samples are filtered according to the maximum sequence length supported by the pre-trained model.

\begin{table}[h]
\centering
\small
\caption{\label{tab:baseline_image_captioning_results} 
BAI augmented training results compared to the baselines, without BAI, on the MS-COCO test set. Metrics are denoted by M=METEOR, R=ROUGE, B=BLEU, S=SPICE and C=CIDEr-D. $\Delta$C denotes the difference in the main score CIDEr-D compared to the baseline "XE only" training.
}
 \begin{tabular}{ | l | l | l | l | l | l | l | l | l | l | l | }
 \hline
Model & Training & B1 & B2 & B3 & B4 & 
R & M & S & C & $\Delta$C \\ 
\hline
Transformer \cite{vaswani2017attention} &  XE & 76.0 & 59.9 & 46.3 & 35.6 & 57.4 & 29.0 & 22.2 & 119.4 & 0.0 \\
Transformer &  XE+BAI   & 76.4 & 60.6 & 47.0 & 36.2 & 57.8 & 29.3 & 22.5 & 120.6 & +1.2 \\
ExpansionNet v2 \cite{hu2023exploiting} &  XE  & 77.6 & 62.0 & 48.2 & 37.2 & 58.1 & 29.4 & 22.5 & 123.5 & 0.0 \\
ExpansionNet v2 &  XE+BAI & 78.2 & 63.0 & 49.2 & 38.1 & 58.5 & 29.5 & 22.8 & 125.9 & +2.4 \\
\hline
\end{tabular}
\end{table}

\begin{table}[h]
\centering
\small
\caption{\label{tab:baseline_ts_results} 
BAI augmented training results compared to the baselines in TS test datasets.  $\Delta$R denotes the difference in the main score ROUGE compared to the baseline (XE). All models are pre-trained according to the modalities of the respective papers and then fine-tuned on the benchmark dataset.
}
 \begin{tabular}{| l | l | l | l | l | l | l |}
\hline
Dataset & Model & Training & BLEU & ROUGE & $\Delta$R \\ 
\hline
TIFU & Flan-T5-Small \cite{chung2022scaling} &  XE  & 20.56 & 32.29 & 0.0 \\
TIFU & Flan-T5-Small & XE+BAI & 21.96 & 33.45 & +1.16 \\
DialogSum & Flan-T5-Small \cite{chung2022scaling} &  XE  & 46.38 & 33.93 & 0.0 \\
DialogSum & Flan-T5-Small & XE+BAI & 46.21 & 34.81 & +0.88 \\
DialogSum & GPT-2 \cite{radford2019language} &  XE  & 39.99 & 24.74 & 0.0 \\
DialogSum & GPT-2 & XE+BAI  & 39.50 & 25.09 & +0.35 \\
\hline
\end{tabular}
\end{table}

\subsection{Training and Models Details}
\label{sec:training_and_models}

The Transformer follows the architecture of the Base Transformer reported in \cite{vaswani2017attention} consisting of $N=6$ encoder and decoder layers, dimension $H=512$, FeedForward size of $FF=2048$ and $8$ attention heads. ExpansionNet v2 follows the configuration of \cite{hu2023exploiting}, it consists of the Swin-Transformer-Large backbone \cite{liu2021swin} and $N=3$ encoder and decoder layers. We adopt the Static expansion coefficients of $G=\{32,64,128,256,512\}$ in the encoder and a dynamic expansion coefficient of 16 in the decoder. The remaining hyper-parameters are the same as the Transformer. The Transformer will be adopted in the case of NMT and TS. ExpansionNet v2 will be deployed for IC. We refer to the original paper for the architecture details of Flan-T5-Small \cite{chung2022scaling}, mBART \cite{liu2020multilingual} and GPT-2 \cite{radford2019language}.

For all three problems, the standard Cross-Entropy (CE) loss is adopted. As described in Equation \ref{eq:bai_augmented_xe} we jointly train CE and BAI using a parameter $\lambda$ which weights the contribution of the BAI loss. 
For each task, we report the batch size, batching criteria, optimizer, learning rate strategy, and the hyper-parameters of the function $\Lambda: t \rightarrow [0, 1] \subset \mathbb{R}$ which provides the BAI weight $\lambda=\Lambda(t)$ in the iteration $t$. We define $\Lambda$ as follows:
\begin{equation}
    \Lambda(t) = \eta + \frac{1}{1 + e^{-(t/T - \phi) / \gamma }} (1 - \eta)
\label{eq:lambda_function}
\end{equation}
where $T$ denotes the number of iterations in one epoch, $\eta$ denote the minimum weight, $\gamma$ determines the velocity of the weight increase and $\phi$ controls the base of the slope.

\textbf{IC Training.} RAdam \cite{liu2019variance} optimizer with $\beta_1=0.9, \beta_2=0.98$, batch size of 48, random batching criteria, an initial learning rate of 2e-4, warmed up for 10000 iterations, then annealed by 0.8 every 2 epochs for 20 epochs. $\Lambda$ is defined by $\eta$=1e-3, $\gamma$=0.5 and $\phi$=15 in the case of ExpansionNet v2. When reported, in the case of the Transformer is configured as $\eta$=1e-7, $\gamma$=2.0 and $\phi$=24.


\textbf{NMT Training.} Adam optimizer \cite{kingma2017adam} with $\beta_1=0.9$ and $\beta_2=0.98$, batching criteria based on the source sentence length, token batch size of 4096 and the Noam learning rate as described in \cite{vaswani2017attention} with 4000 warm-up steps, trained for 300 epochs. The BAI weight function $\Lambda$ is characterized by $\eta$=1e-12,  $\gamma$=0.1 and $\phi$=250 for models trained from scratch. Pre-trained models are fine-tuned for 3 epochs and $\phi$ is set to two.

\textbf{TS Training.} The same optimizer and learning rate for NMT is adopted. Trained for 2 epochs in TIFU and 10 epochs in DialogSum. A batch size of 2048 and no warming steps. $\Lambda$ is defined by $\eta$=1e-4, $\gamma$=0.5 and $\phi$=0.5 for Flan-T5-Small. Whereas $\eta$=1e-12, $\gamma$=0.1 and $\phi$=12 are the configurations for GPT-2.


While the general learning rate of the training follows the convention of the original papers, the definition of the $\Lambda$ function was chosen according to empirical rules formulated during the experimental stage. More details can be found in Section \ref{sec:bai_weight_term_impact}.

\subsection{Evaluation Methods}

In IC, the model is evaluated on the standard evaluation metrics of CIDEr-D 
, METEOR 
, SPICE 
, ROUGE 
,  BLEU 
. In TS we adopt the BLEU and ROUGE scores. For NMT we use the BLEU score only. Beam Search is adopted during inference, with a beam width of 3 in case of IC, and 4 in the case of NMT and TS. 
In the case of pre-trained models, greedy decoding is performed.


\section{Results}

In this Section, we observe the effectiveness of BAI according to different tasks and architectures. We first present the impact of BAI in the Transformer and ExpansionNet v2 when compared with the standard Cross-Entropy training and some related works. 
Then, we will showcase additional analysis to assess the impact of BAI besides standard benchmark metrics. Following, we discuss the impact of the importance of the weight term in the result of BAI. Finally, we showcase some qualitative results.



\subsection{BAI Results}
\label{sec:bai_results}





Tables \ref{tab:baseline_image_captioning_results},  \ref{tab:baseline_ts_results}, and \ref{tab:baseline_nmt_results} report the performance improvements generated by our proposed method against the baselines, in IC, NMT, and TS.

Tables \ref{tab:baseline_image_captioning_results} and \ref{tab:baseline_ts_results} showcase the performance improvements across standard Image Captioning metrics and Text Summarization. It can be observed that, both Transformer and ExpansionNet v2 benefit from our method, with an increase of 1.2 CIDEr-D in the first, and 2.4 CIDEr-D in the latter. Whereas, Flan-T5-Small and GPT-2 reported an increase of 0.88 and 0.35 ROUGE compared to the respective baselines. These results indicate the robustness and flexibility of our approach to different architectures, in particular, the improvement can be seen regardless of the model and pivot selection. In Table \ref{tab:baseline_image_captioning_results} a difference in the improvements of 1.2 CIDEr-D can be observed between the two architectures. This suggests that the method is sensible to the architecture or a proper selection of pivots. However, this characteristic can be regarded as a strength since BAI proved to be effective in the case of the most popular Seq2Seq model, the Transformer, but new architectures and strategies can be developed to benefit more from our approach. 

In Table \ref{tab:baseline_nmt_results} we report the difference in BLEU score for the NMT task. It can be observed that augmenting the Cross-Entropy with BAI is beneficial in both situations, when the model is trained from scratch, represented by the base Transformer, and when it is done during the fine-tuning of a multi-lingual large pre-trained model mBART,  when the standard Cross-Entropy is augmented with BAI. This aspect is also supported by Table
\ref{tab:baseline_ts_results}, which showcase the performance improvements also in the case of the pre-trained models Flan-T5-Small and GPT-2.

\begin{table}[h]
\centering
\caption{ 
\label{tab:baseline_nmt_results}
BAI augmented training results on the IWSLT15 En-Vi test set. $\Delta$BLEU denotes the difference compared to the respective Cross-Entropy-only result. The star symbol $^{\star}$ denotes the model is pre-trained and fine-tuned, otherwise the model is trained from scratch.
}
 \begin{tabular}{| l | l | l | l |}
\hline
Model & Training & BLEU & $\Delta$BLEU \\ 
\hline
Transformer \cite{vaswani2017attention} &  XE & 31.13 & 0.0 \\      
Transformer &  XE+BAI  & 31.62 & +0.49 \\
mBART$^{\star}$ \cite{liu2020multilingual} &  XE  & 44.37 & 0.0 \\      
mBART$^{\star}$ &  XE+BAI  & 49.33 & +4.96 \\
\hline
\end{tabular}
\end{table}

In Table \ref{tab:baseline_nmt_results} a significant difference in the improvements can be noted in the pre-trained model compared to the one trained from scratch. We hypothesize this is caused by the fact that the BAI training appears to be the most effective when the weight term is low at the early stages of the training but high at the final epochs. This situation is perfectly recreated by the pre-training and fine-tuning of mBART, where the BAI's contribution is virtually zero during the pre-training, and it is introduced only when the model reaches the performance plateau. More details can be found in Section \ref{sec:bai_weight_term_impact}.

\begin{table}[h]
\centering
\caption{\label{tab:other_bidir_methods} Impact of different training strategies using L2R and R2L data and the comparison against BAI.}
\begin{tabular}{| l | l | l | l | l | l | l |}
\hline
Training method & B1 & B4 & R & M & S & C \\
\hline
L2R XE (Baseline) & 77.1 & 36.6 & 58.0 & 29.4 & 22.6 & 122.7 \\
R2L XE  & 77.0 & 36.0 & 57.7 & 29.0 & 20.9 & 122.5 \\
R2L XE then L2R XE  & 75.4 & 35.2 & 57.2 & 22.5 & 29.3 & 120.1 \\ 
L2R+R2L XE & 76.0  & 35.5 & 57.3 & 22.4 & 29.1 & 119.3 \\ 
\hline
L2R XE + BAI & 78.2 & 38.1 & 58.5 & 29.5 & 22.8 & 125.9 \\
R2L XE + BAI & 78.0 & 37.7 & 58.1 & 29.3 & 22.8 & 124.3 \\
\hline
\end{tabular}
\end{table}

\subsection{BAI Against Other Bidirectionality Approaches}

While early works on bidirectionality focused on recurrent architectures, in the last years most techniques were developed for the Transformer in different applications. In this regard, we compare our proposal to NAT \cite{gu2017non} (NMT), and CBTIC \cite{zhou2022compact} (for IC) as two representatives of the algorithmic-based work on bidirectional and architecture-based approach for bidirectionality respectively. We train our models using the configurations reported in Section \ref{sec:training_and_models}.

For completeness, we first showcase the ineffectiveness of inducing both Left-to-Right (L2R) and Right-to-Left (R2L) knowledge in the model in simple ways, i.e. without architectural or algorithmic incentives. For example, we use the ExpansionNet architecture for the image captioning task and adopt the training configuration described in Section \ref{sec:training_and_models}. In Table \ref{tab:other_bidir_methods} we observe that combining L2R and R2L data with simple strategies does not lead to improvements and sometimes can even harm the performances. In contrast, BAI improves the output quality regardless of the L2R or R2L\footnote{Training on R2L data leads to artefacts described in \cite{hu2023sacreeos} regardless of BAI.} approach training with an increase of 3.2 and 1.8 CIDEr-D in the two cases.

In Table \ref{tab:other_bidir_methods_updown_ic}, we compare our method against the CBTIC architecture. In particular, we re-create the CBTIC environment by training captioning models on Faster-RCNN features as described in \cite{anderson2018bottom}. Both the architectures of Transformer and ExpansionNet v2, shortened respectively as "Transf." and "ExpV2", significantly benefit from BAI, confirming its robustness to different visual features. In particular, in the "ExpV2" case, BAI can increase the initial score to 3.4 CIDEr compared to the 2.7 increase observed in the CBTIC architecture. Whereas, in the case of Transformer, BAI led to an increase of 17.8. However, the magnitude of such an improvement in the latter case is an outlier that should not be attributed to BAI alone, but rather an unfavourable experimental training setup for the baseline, since it was designed for ExpansionNet v2. 

Compared to CBTIC, BAI improves performance without requiring architectural changes, which is reflected in the constant amount of FLOPs with or without BAI. This aspect can be particularly appreciated in the case of pre-trained models, where architectural modifications can be very time-consuming since they would require the re-training of the foundation model.

In Table \ref{tab:other_bidir_nat_en_de}, we showcase the comparison between our proposal and NAT, in the case of NMT. We re-create the experimental setup of the latter by training from scratch the Transformer architecture presented in Section \ref{sec:training_and_models} on the IWSLT16 En-De dataset. Here, it can be observed that our method improves the BLEU score by 0.3 without changing the model inference speed whereas, NAT, in the vanilla formulation, focuses on reducing the inference time at the cost of slight degradation of performances. 

\begin{table}[h]
\centering
\caption{\label{tab:other_bidir_methods_updown_ic} Comparison between BAI and CBTIC performances on the MS-COCO validation set. Models are trained with Cross-Entropy and Up-Down features \cite{anderson2018bottom}. $\delta$ denotes the CIDEr-D improvement against the respective baselines. FLOP computation assumes a caption length of 20 and 36 visual features include the backbone.}
 \begin{tabular}{| l | l | l | l | l | l |}
\hline
Method & B4 & R & M & C ($\delta$) & FLOPs \\
\hline
ExpV2 & 34.5 & 56.2 & 28.0 & 112.9 (0.0) 
& 8.8 G \\
ExpV2 w\// BAI & 35.3 & 56.8 & 28.2 & 116.3 (+3.4) 
& 8.8 G \\
Transf. & 28.1 & 51.6 & 25.1 & 95.7 
(0.0) 
& 7.63 G \\
Transf. w\// BAI & 34.8 & 56.6 & 28.3 & 113.5 (+17.8) 
& 7.63 G  \\
\hline
Transf. \cite{zhou2022compact} & 35.4 & 56.3 & 27.8 & 111.7 (0.0) 
& 7.63 G \\
CBTIC \cite{zhou2022compact} & 35.6 & 56.8 & 28.1 & 114.4 (+2.7) 
& 8.97 G \\
\hline
\end{tabular}
\end{table}


\begin{table}[h]
\centering
\small
\caption{\label{tab:other_bidir_nat_en_de} Comparison between BAI and NAT in the IWSLT16 En-De. $\delta$ denotes the difference compared to the respective baseline value. NVIDIA Tesla P100 was used in \cite{gu2017non}. Our experiments are performed on NVIDIA A100.}
 \begin{tabular}{| l | l | l |}
\hline
Method & BLEU $(\delta)$ & Latency \\
\hline
Transformer & 29.83 (0.0) & 71 ms \\
Transformer w\// BAI & 30.13 (+0.3) & 71 ms \\
\hline
Transformer \cite{zhou2022compact} & 29.70 (0.0) & 607 ms \\
NAT \cite{zhou2022compact} & 28.16 (-1.54) & 257 ms  \\
\hline
\end{tabular}
\end{table}

\subsection{BAI Weight Term Impact}
\label{sec:bai_weight_term_impact}
 
In this Section, we showcase the impact of the Weight Term choice on the effectiveness of BAI. Although we focused on the case of Image Captioning as the application example for the analysis, similar behaviours were observed in the other selected tasks.

\begin{figure}[h]
  \centering
  \includegraphics[width=1.04\textwidth]{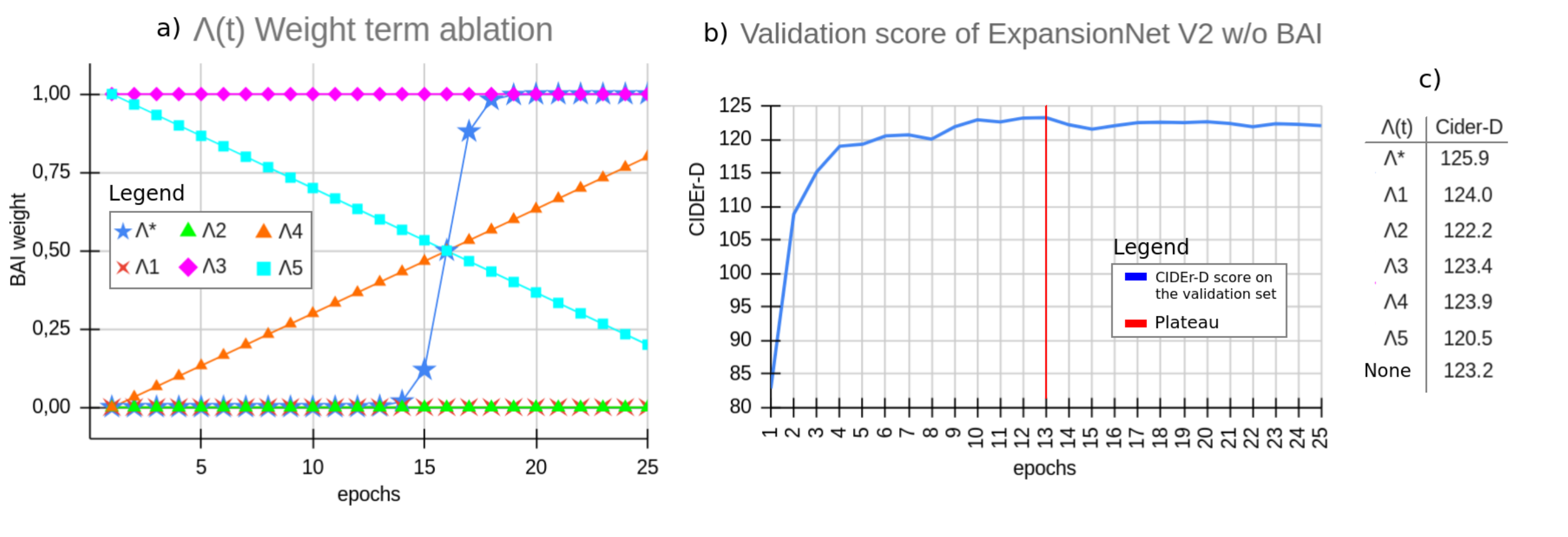}
  \caption{ \label{figure:lr_alternatives} BAI weight function impact on Image Captioning performances. a) Depiction of several strategies of weight term functions. $\Lambda1(t)$=$10^{-3}$, $\Lambda2(t)$=$10^{-6}$, $\Lambda3(t)$=$1$,  $\Lambda4(t)$=$10^{-6}$ $+$  $(1-10^{-6})$ $\cdot$ $(t/30)$, $\Lambda5(t)$=$(1-10^{-6})$ $\cdot$ $(30-t)/30$ $+$ $10^{-6}$, and $\Lambda^*(t)$ is the function described in Section \ref{sec:training_and_models}. $t$ denotes the epoch. b) Validation score of the baseline architecture (ExpansionNet v2) without BAI. The red line denotes the point where the model achieves the highest score. c) Best CIDEr-D score observed from different selections of $\Lambda$.}
\end{figure}

In Figure \ref{figure:lr_alternatives} we plot the weight term function defined in Section \ref{sec:training_and_models} compared to other hand-crafted functions. The alternative weight functions were designed to represent orthogonal criteria such as constant weight, increasing weight, and decreasing weight. 

In Figure \ref{figure:lr_alternatives}.a-c) it can be observed that BAI leads to different results according to the weight term function. For instance, a large term that decreases over time ($\Lambda$5) seems to introduce too much noise to the standard Cross-Entropy learning. However, the problem does not lie in the initial magnitude of the term, as can be seen in the contrary case, $\Lambda$4, where the term increases over the epochs and improves the final CIDEr-D score. In general, the magnitude alone does not suggest a predictable pattern, since the cases of $\Lambda$1, $\Lambda$2, and $\Lambda$3 produced mixed results. Overall, the effectiveness of BAI depends on the weight function and it can be beneficial or detrimental depending on the design of the latter. Fortunately, in the case of $\Lambda^{*}$, we observed that the practice of keeping the term small in the early stages and increasing its magnitude when the model reaches the plateau in the Cross-Entropy loss (Figure \ref{figure:lr_alternatives}-b) empirically leads to the best result. This motivated the design of the weight term function for all three tasks in Section \ref{sec:bai_results}. 

\section{Conclusion and Future Works}

In this work, we tackled the problem of integrating bidirectional awareness into Seq2Seq auto-regressive models. To do so, we first introduce the concept of pivots, defined as network elements that can be trained on auxiliary losses that do not necessarily correlate to the one required by the task but can be helpful in a better accomplishment of the latter. Leveraging this concept, we introduce the Bidirectional Awareness Induction (BAI) and train pivots over a bidirectional loss without breaking the auto-regressive property. The practice appears to increase the quality of the intermediate representations and experimental results involving three architectures, the Transformer, ExpansionNet v2 and GIT and three tasks, such as  Neural Machine Translation, Text Summarization and Image Captioning, showcase our method's robustness, effectiveness, and flexibility. 
Future experiments will focus on addressing the current limitations and explore the training of pivots with additional auxiliary losses, beyond the bidirectional one proposed in this work.

%
%
%
\bibliographystyle{splncs04}
\bibliography{mybibfile}

\end{document}